\documentclass[11pt,a4paper]{article}
\usepackage[hyperref]{emnlp-ijcnlp-2019}
\usepackage{times}
\usepackage{latexsym}
\usepackage{graphicx}
\usepackage{subcaption}
\usepackage{comment}
\usepackage{amsmath}
\usepackage{url}
\usepackage{soul}
\usepackage{todonotes}
\usepackage[normalem]{ulem}
\usepackage{multirow}

\aclfinalcopy % Uncomment this line for the final submission

\newcommand{\qiang}[1]{\textcolor{black}{\small [QN: #1]}}
\newcommand{\qiangchange}[1]{{\color{black} {#1}}}

\newcommand{\SideNote}[2]{\todo[color=#1,size=\small]{#2}} % 
\newcommand{\NoteNP}[1]{\SideNote{purple!40}{#1 --Nanyun}}

\newcommand{\npsuggestion}[1]{\textcolor{purple}{#1}}

\newcommand{\temprel}[1]{\MakeUppercase{\textit{#1}}}
\newcommand{\event}[1]{\textbf{#1}}

\title{Joint Event and Temporal Relation Extraction with Shared Representations and Structured Prediction\\
}

\author{Rujun Han,$^{1,2}$ Qiang Ning,$^3$ Nanyun Peng$^{1,2}$ \\
  $^1$Department of Computer Science, University of Southern California \\
  $^2$Information Sciences Institute, University of Southern California \\
  $^3$Department of Electrical and Computer Engineering, University of Illinois at Urbana-Champaign \\
  {\tt rujunhan@isi.edu, qning2@illinois.edu, npeng@isi.edu}}

\begin{document}
\maketitle
\begin{abstract}
% \qiang{I'm trying to shorten it a bit.}

    We propose a joint event and temporal relation extraction model with shared representation learning and structured prediction.
    The proposed method has two advantages over existing work. First, it improves event representation by allowing the event and relation modules to share the same contextualized embeddings and neural representation learner. Second, it avoids error propagation in the conventional pipeline systems by leveraging structured inference and learning methods to assign both the event labels and the temporal relation labels jointly.
    % (e.g., relations only exist between events and temporal relations cannot form a loop). \RJNote{do we need this example here? the improvement over transitivity is very marginal. event-relation constraint (structure) reflects more of the point of ``joint''. Also, it's described in detail in Intro}
    Experiments show that the proposed method can improve both event extraction and temporal relation extraction over state-of-the-art systems, with the end-to-end $F_1$ improved by 10\% and 6.8\% on two benchmark datasets respectively.

\end{abstract}

\section{Introduction}
%\NoteNPIL{Suggested rearrangement (the current write up mostly followed the flow, but I just want to be clearer).\\
%P1. Task definition and how do existing works cope with the problem.\\
%P2. Why they are not good enough (motivation of why we want to propose our joint model).\\
%P3. Challenges of the new formulation. (here you can bring in~\newcite{miwa-bansal-2016-end} or other joint model and say why we are different and better. Or you can leave~\newcite{miwa-bansal-2016-end} for related work section.}

\begin{figure}[t]
\centering
    \begin{subfigure}[b]{\columnwidth}
    \includegraphics[angle=-90,clip,trim=9cm 2.5cm 7cm 2.5cm, width=\columnwidth]{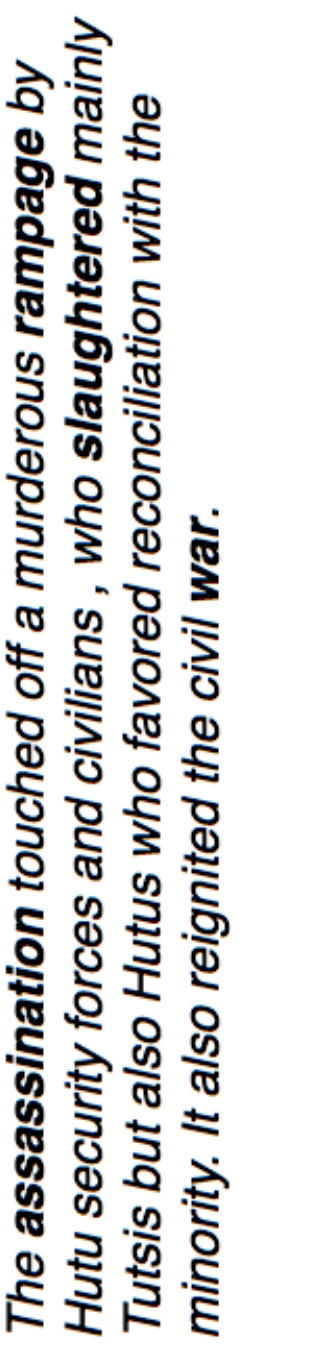}
    \vspace{-0.5cm}
    \end{subfigure}
    
    \begin{subfigure}[b]{0.8\columnwidth}
    \centering
    \includegraphics[angle=-90, clip,trim=3cm 2.2cm 3cm 2.2cm, width=\columnwidth]{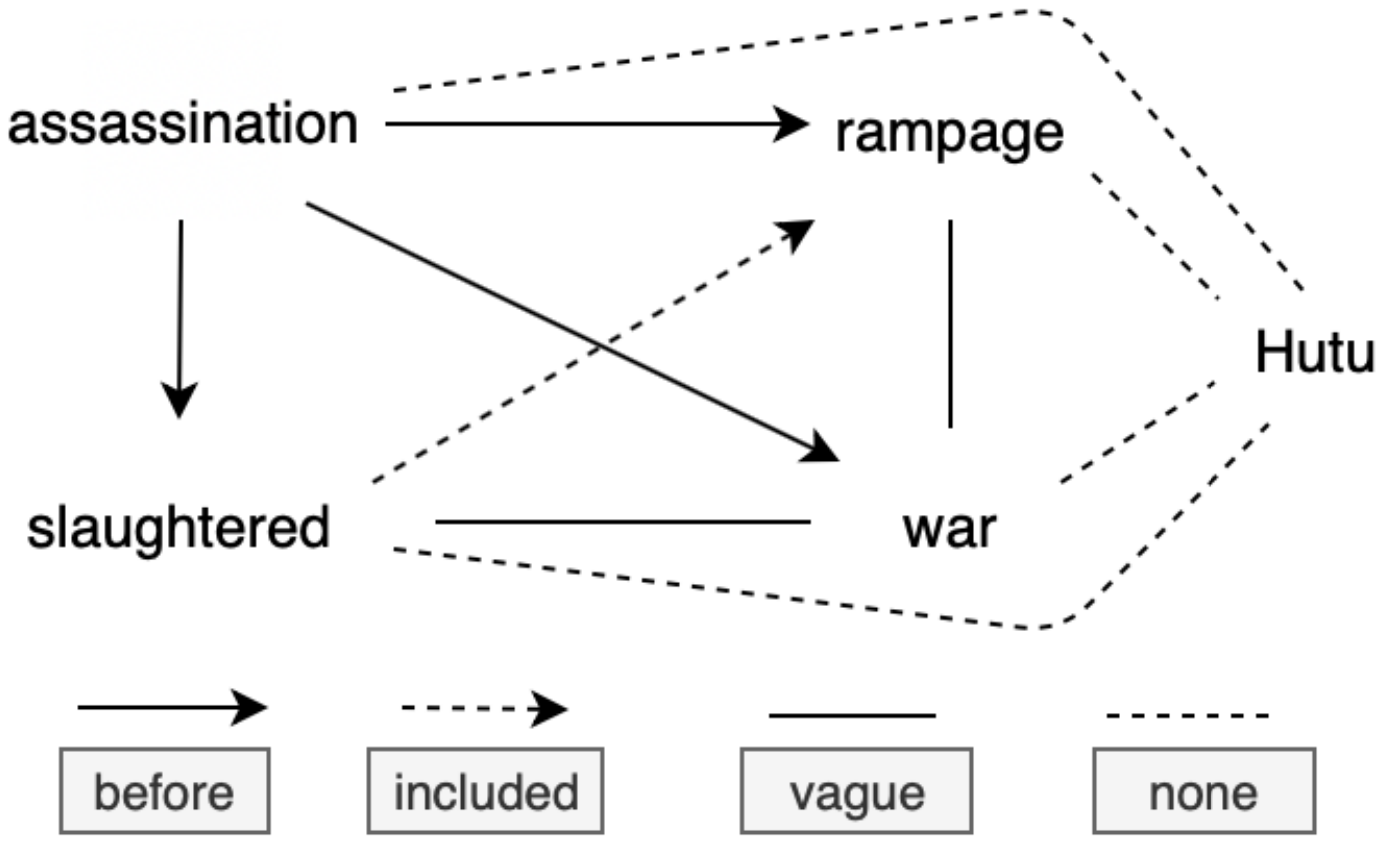}
    \caption{Temporal Relation Graph}
    \label{fig:ex11}
    \end{subfigure}
    
    \begin{subfigure}[b]{0.98\columnwidth}
    \includegraphics[angle=-90, clip, trim=9cm 0cm 9cm 0.5cm, width=0.98\columnwidth]{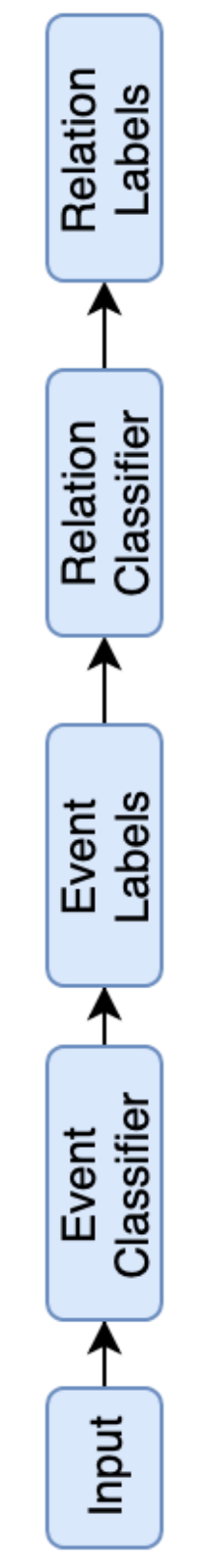}
    \caption{Pipeline Model} 
    \label{fig:ex12}
    \end{subfigure} 
    
    \begin{subfigure}[b]{0.98\columnwidth}
    \includegraphics[angle=-90, clip, trim=6.5cm 0cm 7cm 0.5cm, width=0.98\columnwidth]{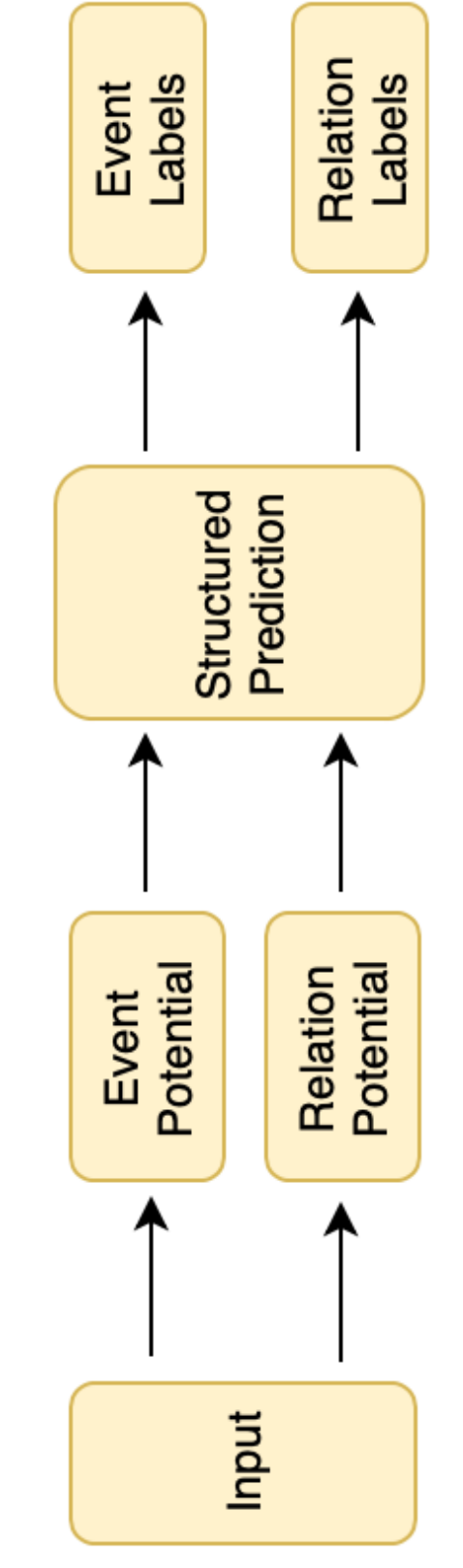}
    \caption{Structured Joint Model} 
    \label{fig:ex13}
    \end{subfigure}

\vspace{-0.1cm}
\caption{An illustration of event and relation models in our proposed joint framework. (a) is a (partial) graph of the output of the relation extraction model. ``Hutu'' is not an event and hence all relations including it should be annotated as \temprel{none}. (b) and (c) are comparisons between \qiangchange{a} pipeline \qiangchange{model} and our joint model.}
\label{fig:ex1} 
\vspace{-2em}
\end{figure}

% \RJNote{Task Definition and previous scope}
%\qiang{I feel that we should start by saying that ``temporal is important'' and then say how the task is defined.}} 
% \NoteNP{how about this: Event temporal relation extraction, which aims at xxx, is a fundamental nlu task.}
The extraction of temporal relations among events 
% aims to understand 
% \RJNote{Just to use a different phrase than aims at} 
% how a story evolves over time. It 
is \qiangchange{an important} natural language understanding (NLU) task that can benefit many downstream tasks such as question answering, information retrieval, and narrative generation. The task \qiangchange{can be modeled as} 
% aims at 
building a graph for a given text, whose nodes represent events
%\qiang{This is the very first sentence and we need to be careful about every word. I think ``entity'' will be confusing and I prefer ``events''. Let me know if you think otherwise.} 
%\RJNote{I am struggling with this too, how about ``event candidate''? Later on in the Method section, I do need to deal with input tokens that are not events} \qiang{Hmm...The aim of the task is definitely to build a graph whose nodes are ``events'', not ``event candidates''. If you agree with that, I think it's clear that we should use ``events'', neither ``entities'' nor ``event candidates''.} 
and edges are labeled with temporal relations correspondingly.
Figure~\ref{fig:ex11} illustrates such a graph 
% an example of such a graph 
for the text shown therein. 
The nodes \event{assassination}, \event{slaughtered}, \event{rampage}, \event{war}, and \event{Hutu} are the candidate events, and different types of edges specify different temporal relations between them: \event{assassination} is \temprel{before} \event{rampage}, \event{rampage} \temprel{includes} \event{slaughtered}, and the relation between \event{slaughtered} and \event{war} is \temprel{vague}. %\RJNote{So we decided to go with Hutu?} \qiang{I don't see any problem with it. Why?}
Since ``Hutu'' is actually not an event, a system is expected to annotate the relations between ``Hutu'' and all other nodes in the graph 
% \qiangchange{(e.g., \event{rampage} and \event{war})}
% , i.e., \event{rampage}, \event{war} etc., 
as \temprel{none} (i.e., no relation).

% \NoteNP{Can we take a real example from the data where our method predict it correctly while the multi-task model make a mistake on the event (and thus the relation)?} 
% \RJNote{I have to push back on this because of both time and space-constraint and not sure how adding another example will help. What makes sense to me is choose a better token say ``forces'' in the above example (which is more likely to be confused as an event than ``Hutu'')}
% \qiang{I agree with RJ on this mainly due to the space issue.}

% \RJNote{Problem of pure pipeline model\qiang{I think this point can be shortened since the problem of a pipeline model is well-accepted. Instead, I think we should work on a short but intuitive explanation to convince people that by looking at a pair of tokens, we get more information to facilitate event identification.}} 
As far as we know, all existing systems treat this task as a pipeline of two separate subtasks, i.e., event extraction and temporal relation classification, and \qiangchange{they also} assume that gold events are given when training the relation classifier ~\cite{Verhagen:2007:STT:1621474.1621488, Verhagen:2010:STT:1859664.1859674, S13-2001, ChambersTBS2014, NingFeRo17, meng2018context}.
%\NoteNP{If we follow Qiang's suggestion, we can do \\cite{ChambersTBS2014} right at the end of the last sentence, and say Figure 1b gives an illustration. However, the errors in event detection naturally propagate into relation model and the signals of relation prediction are not utilized to improve event detection. And talk about the example (a better one than the current figure 1a.)} 
Specifically, they built end-to-end systems 
that extract events \qiangchange{first} and then predict temporal relations between them (Fig.~\ref{fig:ex12}). In these pipeline models, event extraction errors will propagate to the relation classification step and cannot be corrected afterwards. \textbf{Our first contribution} is the proposal of a joint model that extracts both events and temporal relations simultaneously (see Fig.~\ref{fig:ex13}). The motivation is that if we train the relation classifier with \temprel{none} relations between non-events, then it will potentially have the capability of correcting event extraction mistakes. For instance in Fig.~\ref{fig:ex11}, if the relation classifier predicts \temprel{none} for (\event{Hutu}, \event{war}) with a high confidence, then this is a strong signal that can be used by the event classifier to infer that at least one of them is not an event. 
\textbf{Our second contribution} is that we improve event representations by sharing the same contextualized embeddings and neural representation learner between the event extraction and temporal relation extraction modules for the first time. 
On top of the shared embeddings and neural representation learner, the proposed model produces a graph-structured output representing all the events and relations in the given sentences. %also builds individual scoring functions for event and temporal relation extraction. These scores are subsequently used to produce the final graph predictions.

A valid graph prediction in this context should satisfy two structural constraints. First, the temporal relation should always be \temprel{none} between two non-events or between one event and one non-event. Second, for those temporal relations among events, no loops should exist due to the transitive property of time (e.g., if \event{A} is before \event{B} and \event{B} is before \event{C}, then \event{A} must be before \event{C}). 
The validity of a graph is guaranteed by solving an integer linear programming (ILP) optimization problem with those structural constraints, and our joint model is trained by structural support vector machines (SSVM) in an end-to-end fashion.

Results show that, according to the end-to-end $F_1$ score for temporal relation extraction, the proposed method improves CAEVO~\cite{ChambersTBS2014} by 10\% on TB-Dense, and improves CogCompTime~\cite{ning-etal-2018-cogcomptime} by 6.8\% on MATRES. We further show ablation studies to confirm that the proposed joint model with shared representations and structured learning is very effective for this task.

\section{Related Work}
\label{sec:related}

%\qiang{I don't see the necessity of discussing the datasets here because they're actually ``not'' relevant. What's relevant to our contributions are three-fold: how events are extracted, how temporal relations are extracted, and the joint extraction frameworks on other tasks (e.g., entity extraction). So we can postpone the issue of datasets in Sec. 4.}
% \npsuggestion{Our work is closely related to event extraction, relation extraction, and joint models. We briefly summarize \qiangchange{those} related works in this section.}
In this section we briefly summarize the existing work on event extraction and temporal relation extraction.
To the best of our knowledge, there is no prior work on joint \emph{event} and relation extraction, so we will review joint \emph{entity} and relation extraction works instead. 
% \qiang{This is a very good overview for this section.}

% \paragraph{Event Extraction} 
%The series of TempEval competitions \citep{Verhagen:2007:STT:1621474.1621488, Verhagen:2010:STT:1859664.1859674, S13-2001} attract many research interests in predicting event temporal relations.~\npsuggestion{aren't we talking about event here? there are much more event extraction work. you can refer to the related work in this paper~\url{https://www.aclweb.org/anthology/P18-1201}} 
% \RJNote{Just to cite several papers as Violet suggested to complete the stories, not sure if we need all of them as they are not closely related to our event extraction.}
Existing \textbf{event extraction} methods in the temporal relation domain, as in the TempEval3 workshop \citep{S13-2001}, all use conventional machine learning models (logistic regression, SVM, or Max-entropy) with hand-engineered features (e.g., ClearTK \citep{S13-2002} and NavyTime \cite{chambers:2013:SemEval-2013}). While other domains have shown progress on event extraction using neural methods \citep{Nguyen2015EventDA, nguyen-etal-2016-joint, inproceedings}, recent progress in the temporal relation domain is focused more on the setting where gold events are provided. Therefore, we first show the performance of a neural event extractor on this task, although it is not our main contribution.

% Previous event extraction task in other domain leverages either symbolic features \citep{li-etal-2013-joint, Liu2016LeveragingFT} or distributional features \citep{Nguyen2015EventDA, nguyen-etal-2016-joint, inproceedings} with supervised learning method. For temporal event extraction that we focus on, most SOTA event extraction systems such as ClearTK \citep{S13-2002} and NavyTime \cite{chambers:2013:SemEval-2013} are documented in the report of SemEval-2013 Task 1: TEMPEVAL-3 \citep{S13-2001}. These systems commonly adopt simple machine models (Logit, SVM, MaxEnt, etc.) with linguist features and achieve solid performance on the TimeBank dataset.

% \paragraph{Temporal Relation Extraction}
Early attempts on \textbf{temporal relation extraction} use local pair-wise classification with hand-engineered features~\cite{Mani:2006:MLT:1220175.1220270, Verhagen:2007:STT:1621474.1621488, Chambers:2007:CTR:1557769.1557820,Verhagen:2008:TPT:1599288.1599300}. Later efforts, such as ClearTK \citep{S13-2002}, UTTime \citep{laokulrat-EtAl:2013:SemEval-2013}, NavyTime \cite{chambers:2013:SemEval-2013}, and CAEVO \citep{ChambersTBS2014} improve earlier work with better linguistic and syntactic rules.
\citet{yoshikawa2009jointly,NingFeRo17,leeuwenberg2017structured} explore structured learning for this task, and
% based on models with hand-engineered linguistic features. 
more recently, neural methods have also been shown effective~\cite{tourille2017neural, cheng2017classifying, meng2017temporal, meng2018context}.
% which achieve better results than previous linguistic feature-based models \cite{ChambersTBS2014, NingWuRo18}. 
% \newcite{meng2018context} considers global context for pairwise relation predictions on top of the local neural network architecture and achieves the current state-of-the-art (SOTA) performance per micro-average metric.~\NoteNP{I will drop the sota performance bit becasue they only evaluated TBDense, and who is sota on temporal relation extraction isn't important for us.}

% \paragraph{Joint Entity and Relation Extraction} %\NoteNP{Is it fair to claim there isn't really prior joint Event Relation Extraction work? If yes, I'll change this section title to ``joint models for entity and relation extraction''. And say there isn't work for event, so we borrow the work for entity.} %\RJNote{Strictly speaking, previous joint work have tried to tackle entity and relation jointly using global loss function -- see the papers I cite in the introduction and below. I don't think we should claim our work are the first to tackle joint modeling issue.}
%\NoteNPIL{Oh, my focus was not on \emph{joint}, but on \emph{event}. The works you cite here are about joint \emph{entity} and relation extraction, while we are coping with \emph{event} and relation extraction. So I said if there isn't joint \emph{event} and relation work, we'll change the paragraph title to joint \emph{entity} and relation extraction}
In practice, we need to extract both events and those temporal relations among them from raw text. All the works above treat this as two subtasks that are solved in a pipeline. To the best of our knowledge, there has been no existing work on \textbf{joint event-temporal relation extraction}. However, the idea of ``joint'' has been studied for {\em entity}-relation extraction in many works.
% All of the above-mentioned work requires events and relation candidates to be provided, which is not a valid assumption in reality. When deploying a temporal relation extraction system, we need the system to be able to extract both events and relations simultaneously. We are not aware of any prior work on joint event-relation extraction, but joint entity-relation models have been developed in previous research.~\NoteNP{the beginning of the paragraph is a bit odd. You should directly say that the practical setting is to extract both event and relation, but prior work only do it in pipeline. Then you can introduce the joint event-relation work.}
\citet{miwa-sasaki-2014-modeling} frame their joint model as table filling tasks, map tabular representation into sequential predictions with heuristic rules, and construct global loss to compute the best joint predictions. \citet{li-ji-2014-incremental} define a global structure for joint entity and relation extraction, encode local and global features based on domain and linguistic knowledge. and leverage beam-search to find global optimal assignments for entities and relations. \citet{miwa-bansal-2016-end} leverage LSTM architectures to jointly predict both entity and relations, but fall short on ensuring prediction consistency.
% as mentioned in the introduction. 
\citet{zhang-etal-2017-end} combine the benefits of both neural net and global optimization with beam search. 
% However, the computation complexity grows drastically when beam size surpasses 5. 
Motivated by these works, we propose an end-to-end trainable neural structured support vector machine (neural SSVM) model to \emph{simultaneously} extract events and their relations from text and ensure the global structure via ILP constraints. Next, we will describe in detail our proposed method. 
% We formulate the inference problem as integer linear programming (ILP). %tackle global consistency issue by encoding event-relation consistency into integer linear programming (ILP) and train the model end-to-end with feedback signals from ILP.

\begin{figure}[t]

\includegraphics[angle=-90,clip,trim=0cm 2cm 0cm 2cm,width=\columnwidth]{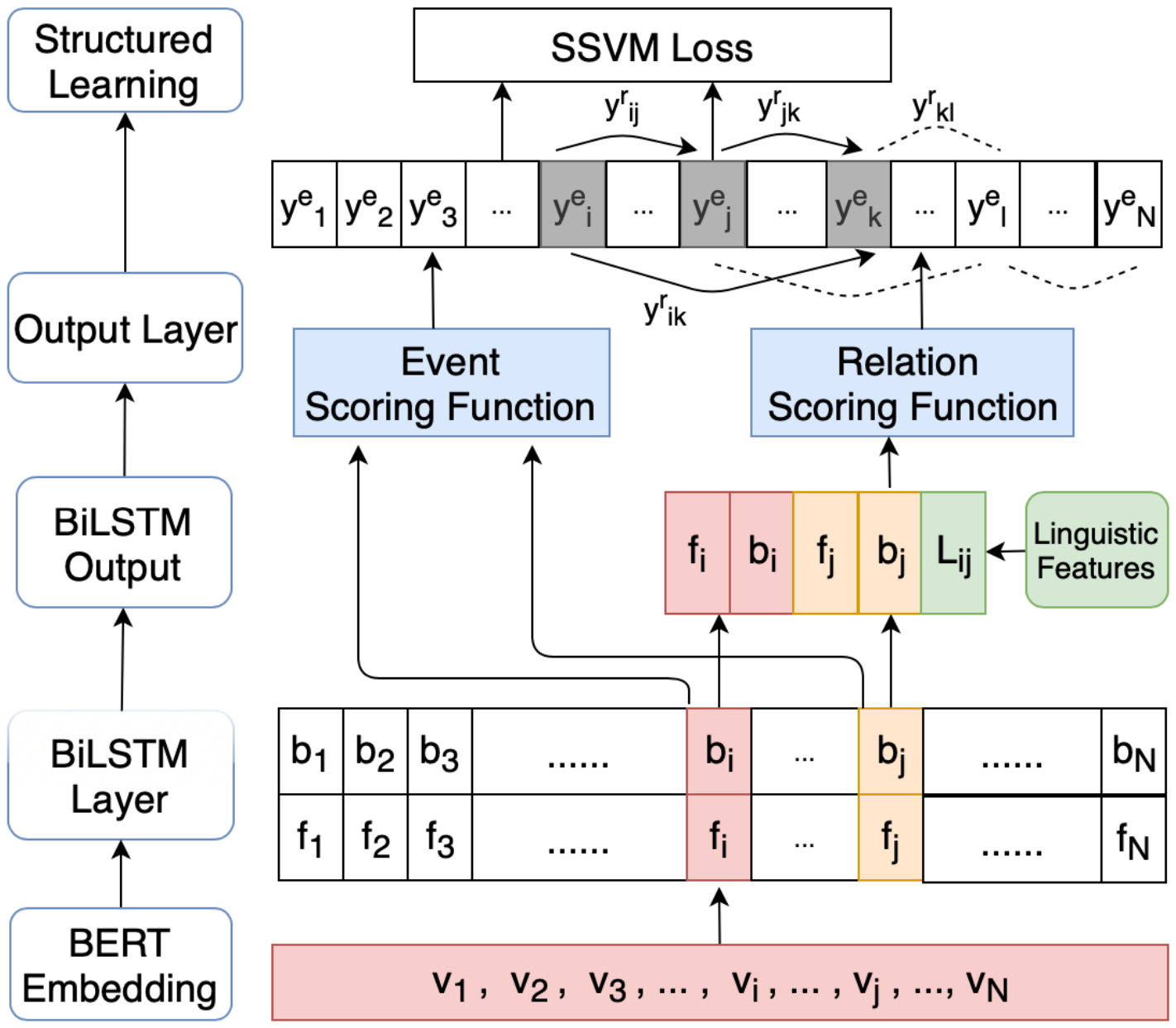}
\caption{
\label{fig:bisltm}
%\qiang{This is not deep, is it? We can simply call ``neural network structure''} 
Deep neural network architecture for joint structured learning. Note that on the structured learning layer, grey bars denote tokens being predicted as events. Edge types between events follow the same notations as in \ref{fig:ex11}. $y^e_l = 0$ (non-event), so all edges connecting \qiangchange{to} $y^e_l$ are \temprel{none}. $y^e_i = 1, y^e_j = 1, y^e_k =1$ (events) and hence edges between them are forced to be the same ($y^r_{ij} = y^r_{jk} = y^r_{ik} =$ \temprel{before} in this example) by transitivity. These global assignments are input to compute the SSVM loss.} %\qiang{``local'' is not very clear; how about ``for event and relation scoring functions''?}
%\hl{I think it's deep since any RNN is deep (N layers). RJ, Please do not forget to added the structured output layer. -NP} }
\vspace{-1.2em}
\end{figure}  

\section{Joint Event-Relation Extraction Model} \label{sec:method}
%\qiang{It's very standard and by no means wrong to name such a section by ``method'', but I feel it's now more popular to give a more informational name here. Say, an acronym of our system, or something like ``joint extraction of events and temporal relations'', which will highlight the contributions of the work.}
\vspace{-0.5em}
In this section we first provide an overview of our neural SSVM model, and then describe each component in our framework in detail (i.e., \qiangchange{the} multi-tasking neural scoring module, \qiangchange{and how inference and learning are performed}).
We denote the set of all possible relation labels (including \temprel{none}) as $\mathcal{R}$, all event candidates (both events and non-events) as $\mathcal{E}$, and all relation candidates as $\mathcal{E}\mathcal{E}$. %\RJNote{nope, we can have negative relations}
\vspace{-0.5em}
\subsection{Neural SSVM} %\NoteNP{I'd like to formulate this as an unified loss function with entity potential/factor and relation potential/factor, instead of an entity loss and a relation loss.}
%In our event temporal relation prediction task, traditional structured support vector machine loss can be modified as,
%\RJNote{Please see if the following loss function formula make sense to you}
Our neural SSVM adapts the SSVM loss as: 
\vspace{-.5em}
{
\begin{align} \label{eq:loss}
\mathcal{L} = & \sum_{n = 1}^l \frac{C}{M^n} \big[\max_{\hat{\boldsymbol{y}}^n \in \mathcal{Y}} \big(0, \Delta(\boldsymbol{y}^n, \hat{\boldsymbol{y}}^n) \text{ } + \bar{S}^n_{\mathcal{R}} \\ &
+ C_{\mathcal{E}}\bar{S}^n_{\mathcal{E}} \big)\big] + ||\Phi||^2, \nonumber
\end{align}
}
\qiangchange{where}
$\bar{S}^n_{\mathcal{E}} = S(\hat{\boldsymbol{y}}^n_\mathcal{E}; \boldsymbol{x}^n) - S(\boldsymbol{y}^n_\mathcal{E};\boldsymbol{x}^n)$ and
$\bar{S}^n_{\mathcal{R}} = S(\hat{\boldsymbol{y}}^n_\mathcal{R}; \boldsymbol{x}^n) - S(\boldsymbol{y}^n_\mathcal{R};\boldsymbol{x}^n)$\qiangchange{;}
$\Phi$ denotes model parameters, $n$ indexes instances, $M^n = |\mathcal{E}|^n + |\mathcal{E}\mathcal{E}|^n$ denotes the total number of relations $|\mathcal{E}|^n$ and events $|\mathcal{E}\mathcal{E}|^n$ in instance $n$. $\boldsymbol{y}^n,\hat{\boldsymbol{y}}^n$ denote the gold and predicted global assignments of events and relations for instance $n$---each of which consists of either one hot vector representing true and predicted relation labels $\boldsymbol{y}_{\mathcal{R}}^n, \boldsymbol{\hat{y}}_{\mathcal{R}}^n \in \{0, 1\}^{|\mathcal{E}\mathcal{E}|}$, or entity labels $\boldsymbol{y}_{\mathcal{E}}^n, \boldsymbol{\hat{y}}_{\mathcal{E}}^n \in \{0, 1\}^{|\mathcal{E}|}$. 
A maximum a posteriori probability (MAP) inference is needed to find $\boldsymbol{\hat{y}}^n$, which we formulate as an interger linear programming (ILP) problem and describe more details in Section~\ref{sec:inference}.
$\Delta(\boldsymbol{y^n}, \hat{\boldsymbol{y}}^n)$ is a distance measurement between the gold and the predicted assignments; we simply use the Hamming distance.  $C$ and $C_{\mathcal{E}}$ are the hyper-parameters to balance the losses between event, relation and the regularizer, and $S(\boldsymbol{y}^n_\mathcal{E};\boldsymbol{x}^n), S(\boldsymbol{y}^n_\mathcal{R};\boldsymbol{x}^n)$ are scoring functions, which we design a multi-tasking neural architecture to learn. % (more details in the following sections) %simply use a constant if $\boldsymbol{y} \neq \hat{\boldsymbol{y}}$, and $0$ otherwise

The intuition behind the SSVM loss is that it requires the score of gold output structure $\boldsymbol{y}^n$ to be greater than the score of the best output structure under the current model $\hat{\boldsymbol{y}}^n$ with a margin $\Delta(\boldsymbol{y}^n, \hat{\boldsymbol{y}}^n)$\footnote{Note that if the best prediction is the same as the gold structure, the margin is zero; there will be no loss.} or else there will be some loss. The training objective is to minimize the loss.

The major difference between our neural-SSVM and the traditional SSVM model is the scoring function. Traditional SSVM uses a linear function over hand-crafted features to compute the scores, whereas we propose to use a recurrent neural network to estimate the scoring function and train the entire architecture end-to-end. 
%\vspace{-0.2cm}

\subsection{Multi-Tasking Neural Scoring Function}
The recurrent neural network (RNN) architecture has been widely adopted by prior temporal extraction work to encode context information ~\cite{tourille2017neural, cheng2017classifying, meng2017temporal}. Motivated by these works, we adopt a RNN-based scoring function for both event and relation prediction in order to learn features in a data driven way and capture long-term contexts in the input. 
%\NoteNP{We should stop calling it ``the local neural architecure'', also we should probably motivate it by saying RNN is a widely-adopted encoder including prior temporal extraction work.} 
In Fig.~\ref{fig:bisltm}, we skip the input layer for simplicity.\footnote{Following the convention of event relation prediction literature \citep{ChambersTBS2014, P18-1212, NingWuRo18}, we only consider event pairs that occur in the same or neighboring sentences, but the architecture can be easily adapted to the case where inputs are longer than two sentences.} 

The bottom layer corresponds to contextualized word representations denoted as $v_k$. We use ($i, j$) $\in \mathcal{E}\mathcal{E}$ to denote a candidate relation and $i \in \mathcal{E}$ to indicate a candidate event in the input sentences of length N. We fix word embeddings computed by a pre-trained BERT-base model \citep{BERT2018}. They are then fed into a BiLSTM layer to further encode task-specific contextual information. Both event and relation tasks share this layer.
%\qiang{you used BiLSTM in other places, so please be consistent throughout the paper. The same issue also happens for the temporal relation: NONE, None, and {\em None}. For temporal relations, I suggest that you define a macro for it so that they will be consistent throughout the paper}

The event scorer is illustrated by the left two branches following the BiLSTM layer. We simply concatenate both forward and backward hidden vectors to encode the context of each token. As for the relation scorer shown in the right branches, for each pair ($i,j$) we take the forward and backward hidden vectors corresponding to them, $f_i, b_i, f_j, b_j$, and concatenate them with linguistic features as in previous event relation prediction research. We denote linguistic features as $L_{i,j}$ and only use simple features provided in the original datasets: token distance, tense, and polarity of events.

%\NoteNP{Should we mention MLP classifiers now? We only need a softmax layer to produce probabilities, but not a classifier.}
Finally, all hidden vectors and linguistic features are concatenated to form the input to compute the probability of being an event or a softmax distribution over all possible relation labels---which we refer to as the RNN-based \textbf{scoring function} in the following sections.

\subsection{MAP Inference} \label{sec:inference}
A MAP inference is needed both during training to obtain $\hat{\boldsymbol{y}}^n$ in the loss function (Equation~\ref{eq:loss}), as well as during the test time to get globally coherent assignments. 
We formulate the inference problem as an ILP problem. The inference framework is established by constructing a global objective function using scores from local scorers and imposing several global constraints: 1) one-label assignment, 2) event-relation consistency, and 3) symmetry and transitivity as in \citet{BramsenPYR12, ChambersJ2008, DenisM2008, DoLuRo12, NingFeRo17}.

\subsubsection{Objective Function} 
The objective function of the global inference is to find the global assignment that has the highest probability under the current model, as specified in Equation~\ref{eq:infObj}: %maximizes the score of global assignments as specified in Equation~\ref{eq:infObj}\footnote{The objective function is specified on the instance level.}. 

\vspace{-1em}
{\small
\begin{align} \label{eq:infObj}
\hat{y} = & \arg \max  \sum_{(i,j) \in \mathcal{E}\mathcal{E}} \sum_{r \in \mathcal{R}} y^r_{i,j} S(y^r_{i,j},\boldsymbol{x}) \\ & + C_{\mathcal{E}} \sum_{k \in \mathcal{E}} \sum_{e \in \{0, 1\}} y^e_k S(y^e_k,\boldsymbol{x}) \nonumber
\end{align}
\[
\textbf{s.t.      } y^r_{i,j}, y^e_k \in \{0, 1\} \text{  , } \sum_{r \in \mathcal{R}} y^r_{i,j} = 1, \sum_{e \in \{0, 1\}} y^e_k = 1,
\]
}where $y^e_k$ is a binary indicator of whether the $k$-th candidate is an event or not, and $y^r_{i,j}$ is a binary indicator specifying whether the global prediction of the relation between $(i,j)$ is $r \in \mathcal{R}$. $S(y^e_k,\boldsymbol{x}), \forall e \in \{0, 1\}$ and $S(y^r_{i,j},\boldsymbol{x}), \forall r \in \mathcal{R}$ are the scoring functions obtained from the event and relation scoring functions, respectively. The output of the global inference $\bf{\hat{y}}$ is a collection of optimal label assignments for all events and relation candidates in a fixed context. $C_{\mathcal{E}}$ is a hyper-parameter controlling weights between relation and event. The constraint that follows immediately from the objective function is that the global inference should only assign one label for all entities and relations.
%~\NoteNP{you probably want to add this $C_{\mathcal{E}}$ into the loss function in Equation~\ref{eq:loss}} 

\subsubsection{Constraints}
We introduce several additional constraints to ensure the resulting optimal output graph forms a valid and plausible event graph.
\paragraph{Event-Relation Consistency.}
Event and relation prediction consistency is defined with the following property: \textbf{a pair of input tokens have a positive temporal relation if and only if both tokens are events}. The following global constraints will satisfy this property,
\[
\forall (i,j) \in \mathcal{E}\mathcal{E}, e^P_i \ge r^P_{i,j} \text{ , } e^P_j \ge r^P_{i,j} 
\]
\[
\text{and } e^N_i + e^N_j \ge r^N_{i,j}
\]where $e^P_i$ denotes an event and $e^N_i$ denotes a non-event token. $r^P_{i,j}$ indicates positive relations: \temprel{before}, \temprel{after}, \temprel{simultaneous}, \temprel{includes}, \temprel{is\_included}, \temprel{vague} and $r^N_{i,j}$ indicate a negative relation, i.e., \temprel{none}. A formal proof of this property can be found in  Appendix A.

\paragraph{Symmetry and Transitivity Constraint.}
We also explore the symmetry and transitivity constraints of relations. They are specified as follows: %\RJNote{I think the current flow works fine. Showing the equation first and then detailed explanation.}
%~\NoteNP{You need to give more details about what's symmetry and transitivity constraints. (oh, you only need to move the next paragraph up.)} 

{\small
\[
\forall (i,j), (j,k) \in \mathcal{E}\mathcal{E}, y^r_{i,j} = y^{\bar{r}}_{j,i}, \text{ (symmetry)}
\]
\[
y^{r_1}_{i,j} + y^{r_2}_{j,k} - \sum_{r_3 \in Trans(r_1, r_2)} y^{r_3}_{i,k} \leq 1, \text{ (transitivity)}
\]
}

Intuitively, the symmetry constraint forces two pairs of events with flipping orders to have reversed relations. For example, if $r_{i,j}$ = \textit{BEFORE}, then $r_{j,i}$ = \textit{AFTER}. The transitivity constraint rules that if ($i,j$), ($j,k$) and ($i,k$) pairs exist in the graph, the label (relation) prediction of ($i,k$) pair has to fall into the transitivity set specifyed by ($i,j$) and ($j,k$) pairs. The full transitivity table can be found in \citet{P18-1212}.

\subsection{Learning} 
%\NoteNP{You may want to start with saying we can directly optimize Equation~\ref{eq:loss}, but it didn't work well in our preliminary experiments, so we develop the two-stage learning. then direct everything into the implementation details.} \qiang{I agree.}
%\NoteNP{Did you warmed up with training on gold annotations (relations are predicted based on gold events)?}
%\RJNote{explained in implementation details.}

We begin by experimenting with optimizing SSVM loss directly, but model performance degrades.\footnote{We leave further investigation for future work.} Therefore, we develop a two-state learning approach which first trains a pipeline version of the joint model without feedback from global constraints. In other words, the local neural scoring functions are optimized with cross-entropy loss using gold events and relation candidates that are constructed directly from the outputs of the event model. During the second stage, we switch to the global SSVM loss function in Equation~\ref{eq:loss} and re-optimize the network to adjust for global properties. We will provide more details in Section~\ref{sec:implementation}. %Note that the local model is trained with cross-entropy loss, which is replaced with SSVM loss during the second stage.

\section{Implementation Details} \label{sec:implementation}
%~\NoteNP{Baselines and implementation details? Should also talk about the CAVEO and CogCompTimeSingle baselines. }
In this section we describe implementation details of the baselines and our four models to build an end-to-end event temporal relation extraction system with an emphasis on the structured joint model. In Section~\ref{sec:results} we will compare and contrast them and show why our proposed structured joint model works the best.

\subsection{Baselines} 
We run two event and relation extraction systems, CAEVO\footnote{\url{ https://www.usna.edu/Users/cs/nchamber/caevo/}} \citep{ChambersTBS2014} and CogCompTime\footnote{\url{http://cogcomp.org/page/publication_view/844}} \citep{ning-etal-2018-cogcomptime}, on TB-Dense and MATRES, respectively. These two methods both leverage conventional learning algorithms (i.e., MaxEnt and averaged perceptron, respectively) based on manually designed features to obtain separate models for events and temporal relations, and conduct end-to-end relation extraction as a pipeline. 
% \RJNote{Qiang, please correct me if I am wrong}.
% and report its event and end-to-end relation extraction performance as the baseline for TB-Dense. 
Note \citet{ChambersTBS2014} does not report event and end-to-end temporal relation extraction performances, 
% does not report these two numbers, 
so we calculate the scores per our implementation.
%~\NoteNP{friends, we should add more details about the baselines (e.g. feature-based? neural-based? used bert or not?)} 
% \RJNote{Qiang, I feel you are in a better position to concisely summarize CogCompTime baselines than me.}

\subsection{End-to-End Event Temporal Relation Extraction}
\paragraph{Single-Task Model.} The most basic way to build an end-to-end system is to train separate event detection and relation prediction models with gold labels, as we mentioned in our introduction. In other words, the BiLSTM layer is not shared as in Fig.~\ref{fig:bisltm}. During evaluation and test time, we use the outputs from the event detection model to construct relation candidates and apply the relation prediction model to make the final prediction. 
%We implement single-task model for two purposes: 1) to verify that our NN architecture can outperform or achieve comparable performance from previous SOTA systems; 2) to serve as a baseline for the end-to-end task this paper focuses on.

% \NoteNP{We may want to frame the models from here on as the variations of joint models (which haven't been explored in the event and relation extraction context)}
% \RJNote{Not sure what you mean exactly, please provide more details or examples.}
\paragraph{Multi-Task Model.} This is the same as the single-task model except that the BiLSTM layer is now shared for both event and relation tasks. Note that both single-task and multi-task models are not trained to tackle the \temprel{none} relation directly. They both rely on the predictions of the event model to annotate relations as either positive pairs or \temprel{none}.
%\NoteNP{I don't understand from here onward.} 

\paragraph{Pipeline Joint Model.} This shares the same architecture as the multi-task model, except that during training, we use the predictions of the event model to construct relation candidates to train the relation model. This strategy will generate \temprel{none} pairs during training if one argument of the relation candidate is not an event. These \temprel{none} pairs will help the relation model to %are inputs to the relation model so that the model can directly 
distinguish negative relations from positive ones, and thus become more robust to event prediction errors. 
We train this model with gold events and relation candidates during the first several epochs in order to obtain a relatively accurate event model and switch to a pipeline version afterwards inspired by~\citet{miwa-bansal-2016-end}.

\paragraph{Structured Joint Model.} This is described in detail in Section~\ref{sec:method}. However, we experience difficulties in training the model with SSVM loss from scratch. %observe model performance is much lower than the pipeline joint model if we train the model with SSVM loss from scratch. 
This is due to large amounts of non-event tokens, and the model is not capable of distinguishing them in the beginning. We thus adopt a two-stage learning procedure where we take the best pipeline joint model and re-optimize it with the SSVM loss. 

 %\RJNote{Moved justification of Adam vs. SGD in the appendix. I don't think this is crucial as the differential power comes from the global correction using ILP and optimizer will drive the model to a different optimum with new ``training signal''. But I agree it could be misleading if the reviewer doesn't quite get our model framework.}  

To restrict the search space for events in the ILP inference of the SSVM loss, we use the predicted probabilities from the event detection model to filter out non-events since the event model has a strong performance, as shown in Section~\ref{sec:results}. Note that this is very different from the pipeline model where events are first predicted and relations are constructed with \emph{predicted} events. Here, we only leverage an additional hyper-parameter $T_{evt}$ to filter out highly unlikely event candidates. Both event and relation labels are assigned \emph{simutaneously} during the global inference with ILP, as specified in Section~\ref{sec:inference}. We also filter out tokens with POS tags that do not appear in the training set as most of the events are either nouns or verbs in TB-Dense, and all events are verbs in MATRES.
\vspace{-0.3em}
\paragraph{Hyper-Parameters.} All single-task, multi-task and pipeline joint models are trained by minimizing cross-entropy loss. We observe that model performances vary significantly with dropout ratio, hidden layer dimensions of the BiLSTM model and entity weight in the loss function (with relation weight fixed at 1.0). We leverage a pre-trained BERT model to compute word embedding\footnote{We use a pre-trained BERT-Base model with 768 hidden size, 12 layers, 12 heads implemented by \url{https://github.com/huggingface/pytorch-pretrained-BERT}} and all MLP scoring functions have one hidden layer.\footnote{Let $H, K$ denotes the dimension of (concatenated) vector from BiLSTM and number of output classes. MLP layer consists of $|H| * |K| + |K| * |K|$ parameters} In the SSVM loss function, we fix the value of $C = 1$, but fine-tune $C_\mathcal{E}$ in the objective function in Equation~\ref{eq:infObj}.
Hyper-parameters are chosen using a standard development set for TB-Dense and a random holdout-set based on an 80/20 split of training data for MATRES. To solve ILP in the inference process, we leverage an  off-the-shelf solver provided by Gurobi optimizer; i.e. the best solutions from the Gurobi optimizer are inputs to the global training. The best combination of hyper-parameters can be found in Table 9 in our appendix.\footnote{PyTorch code will be made available upon acceptance.}

\section{Experimental Setup}
In this section we first provide a brief overview of temporal relation data and describe the specific datasets used in this paper. We also explain the evaluation metrics at the end. % and provide details regarding our model implementation and experiments.

\subsection{Temporal Relation Data} \label{sec:data}
Temporal relation corpora such as TimeBank \citep{PustejovskyX2003} and RED \citep{O'Gorman2016} facilitate the research in temporal relation extraction. The common issue in these corpora is missing annotations. Collecting densely annotated temporal relation corpora with all events and relations fully annotated is reported to be a challenging task as annotators could easily overlook some facts \citep{Bethard:2007:TTI:1304608.1306306, P14-2082, ChambersTBS2014, NingFeRo17}, which made both modeling and evaluation extremely difficult in previous event temporal relation research. 
%~\NoteNP{I'll remove the moreover bit since people could argue you can conduct unsupervised learning, learning from partial annotation, etc.} 

The TB-Dense dataset mitigates this issue by forcing annotators to examine all pairs of events within the same or neighboring sentences, and it has been widely evaluated on this task \cite{ChambersTBS2014, NingFeRo17, cheng2017classifying, meng2018context}. Recent data construction efforts such as MATRES~\citep{P18-1212} further enhance the data quality by using a multi-axis annotation scheme and adopting a start-point of events to improve inter-annotator agreements.
We use TB-Dense and MATRES in our experiments and briefly summarize the data statistics in Table~\ref{tab:data}. %We focus on event relation in this work and hence all numbers refer to $\mathcal{E}\mathcal{E}$ pairs. %\RJNote{We have two-paragraph as justification of data selection, but just add previous publications}
%~\NoteNP{If possible, we should make sure the overview of the datasets is thorough, and we should motivate a little about our data selection. E.g. very representative datasets with relatively large amount of dense annotations and prior published results as benchmarks.}

%\NoteNP{We can remove these two paragraphs for space if needed.}
%\paragraph{TB-Dense}~\citet{P14-2082} annotates 36 documents in TimeBank corpus and addresses the sparse-annotation issue in the original data by introducing the \temprel{vague} label and requiring annotators to label all pairs of events/times in a given window.

%\paragraph{MATRES}~\citet{NingWuRo18} is based on all TE-3 Workshop corpus, but filters out non-verbal events. TE-3 has three datasets: TimeBank (TB), Aquaint (AQ) and Platinum (PT). TB and AQ are used as training sets and PT is used as test set. The authors project events on multiple axes and only keep those in the main-axis. Start-point temporal scheme is adopted when out-sourcing the annotation task, which contributes to performance improvement of machine learning models built on relation extraction tasks.

 \begin{table}[t]
    \small
    %\smallskip
 	\centering
 	\begin{tabular}{l|c|c} \hline
 	& TB-Dense & MATRES \\ \hline\hline
    \multicolumn{3}{c}{\textbf{\# of Documents}} \\ \hline
 	Train & 22 &  183\\
 	Dev  &  5 &  -\\
    Test & 9 & 20 \\\hline\hline
    \multicolumn{3}{c}{\textbf{\# of Pairs}} \\ \hline
 	Train & 4032 &  6332\\
 	Dev & 629 & - \\
 	Test & 1427 & 827\\\hline
 	\end{tabular}
   	\caption{Data overview. Note that the numbers reported for MATRES do not include the AQUAINT section.}
   	\label{tab:data}
   	\vspace{-0.3cm}
 \end{table}
 
  \begin{table*}
 	\centering
 	\begin{tabular}{|l|l|ccc|ccc|} \hline
 	\multirow{2}{*}{\textbf{Corpus}} & \multirow{2}{*}{\textbf{Models}} &\multicolumn{3}{c|}{\textbf{Event}} & \multicolumn{3}{c|}{\textbf{Relation}}\\ %\cline{3-8}
 	&&\textbf{P}& \textbf{R}&\textbf{F1} & \textbf{P}& \textbf{R} & \textbf{F1}\\ \hline\hline
 	TB-Dense & Structrued Joint Model (Ours) & 89.2 & \textbf{92.6} &\textbf{90.9} & \textbf{52.6} & \textbf{46.5} & \textbf{49.4} \\
 	& \citet{ChambersTBS2014} & \textbf{97.2} & 79.4 & 87.4 & 43.8 & 35.7 & 39.4\\\hline
 	MATRES & Structrued Joint Model (Ours) & \textbf{87.1} & \textbf{88.5} & \textbf{87.8} & \textbf{59.0}& \textbf{60.2}& \textbf{59.6}\\
 	& \citet{ning-etal-2018-cogcomptime} & 83.5 & 87.0 & 85.2 & 48.4&58.0&52.8\\\hline
 	\end{tabular}
   	\caption{Event and Relation Extraction Results on TB-Dense and MATRES}
   	\label{tab:main-perf}
 \end{table*}

  \begin{table*}
 	\centering
 	%\small
 	\setlength{\tabcolsep}{0.5em}
 	\begin{tabular}{|l|c|c|c|c|c|c|} \hline
 	\textbf{Micro-average}&\multicolumn{3}{|c|}{\textbf{TB-Dense}} & \multicolumn{3}{|c|}{\textbf{MATRES}}\\ \cline{2-7}
 	\textbf{F1 (\%)}&\textbf{Event}& \textbf{Relation (G)} & \textbf{Relation (E)} & \textbf{Event}& \textbf{Relation (G)} & \textbf{Relation (E)}\\ \hline\hline
 	Baselines & 87.4$\dag$ & 57.0$\ddag$ & 39.4$\dag$ &  85.2 &  65.9& 52.8\\\hline\hline
 	Single-task & 88.6 & 61.9 & 44.3 &  86.9 &  75.3& 57.2\\
 	Multi-task & 89.2 & \textbf{64.5} & 48.4&  86.4&  \textbf{75.5}& 58.7\\
 	Pipeline Joint &90.3 & - & 48.5& 87.2 & -  &58.5\\
 	Structured Joint &\textbf{90.9} & - & \textbf{49.4} & \textbf{87.8} & - & \textbf{59.6}\\\hline
 	\end{tabular}
   	\caption{Further ablation studies on event and relation extractions. Relation (G) denotes train and evaluate using gold events to compose relation candidates, whereas Relation (E) means end-to-end relation extraction. $\dag$ is the event extraction and pipeline relation extraction F1 scores for CAEVO~\cite{ChambersTBS2014}. 57.0$\ddag$ is the best previously reported micro-average score for temporal relation extraction based on gold events by \citet{meng2018context}. All MATRES baseline results are provided by \citet{ning-etal-2018-cogcomptime}.}
   	\label{tab:overall-perf}
   \vspace{-1em}
 \end{table*}
 
\vspace{-0.8em}
\subsection{Evaluation Metrics}
To be consistent with previous research, we adopt two different evaluation metrics. The first one is the standard \textbf{micro-average scores}. For densely annotated data, the micro-average metric should share the same precision, recall and F1 scores. However, since our joint model includes \temprel{none} pairs, we follow the convention of IE tasks and exclude them from evaluation. The second one is similar except that we exclude both \temprel{none} and \temprel{vague} pairs following~\cite{ning-etal-2018-cogcomptime}. Please refer to Figure 4 in the appendix  for a visualizations of the two metrics.

\section{Results and Analysis}
\label{sec:results}

The main results of this paper can be found in Table~\ref{tab:main-perf}. All best-recall and F1 scores are achieved by our structured joint model, and the results outperform the baseline systems by 10.0\% and 6.8\% on end-to-end relation extraction per F1 scores and 3.5\% and 2.6\% on event extraction per F1 scores. The best precision score for the TB-Dense dataset is achieved by CAEVO, which indicates that the linguistic rule-based system can make highly precise predictions by being conservative.

Table~\ref{tab:overall-perf} shows a more detailed analysis, in which we can see that our single-task models with BERT embeddings and a BiLSTM encoder already outperform the baseline systems on end-to-end relation extraction tasks by 4.9\% and 4.4\% respectively. In the following sections we discuss step-by-step improvement by adopting multi-task, pipeline joint, and structured joint models on end-to-end relation extraction, event extraction, and relation extraction on gold event pairs.

\subsection{End-to-End Relation Extraction}
%~\NoteNP{We suggest to follow the presentation of table 1 and table 2 in this paper \url{https://arxiv.org/pdf/1601.00770.pdf}}
 %\temprel{includes}, \temprel{is\_included} and \temprel{vague} have low scores because of small number of training pairs in the data.

%\NoteNP{why do you need to discuss the \temprel{none} pairs? don't we suppose to compare single task v.s. multi-task, v.s. joint models? Can we somehow follow the flow for MATRES as I just updated?}
\paragraph{TB-Dense.} 
The improvements over the single-task model per F1 score are 4.1\% and 4.2\% for the multi-task and pipeline joint model respectively. This indicates that the pipeline joint model is helpful only marginally. 
Table \ref{tab:tbd-breakdown} shows that the structured joint model improves both precision and recall scores for \temprel{before} and \temprel{after} and achieves the best end-to-end relation extraction performance at 49.4\%---which outperforms the baseline system by 10.0\% and the single-task model by 5.1\%. 

%We introduce \temprel{none} label to deal with pairs that are not qualified for positive temporal relations. Due to large amount of non-event tokens, the number of \temprel{none} pairs could explode if we do not apply any filter to restrict search space. However, even after applying all filters mentioned in Section~\ref{sec:implementation}, we still observe ~15\% negative pairs when evaluating the test data. This explains the large performance decline compared to the evaluation on gold pairs (Table~\ref{tab:overall-perf} Column Relation (G) vs. Relation (E)) as \temprel{none} pairs can be predicted as positive pairs and vice versa. Structured joint model can help address this issue. 

\paragraph{MATRES.} Compared to the  single-task model, the multi-task model improves F1 scores by 1.5\%, while the pipeline joint model improves F1 scores by 1.3\%---which means that pipeline joint training does not bring any gains for MATRES. 
The structured joint model reaches the best end-to-end F1 score at 59.6\%, which outperforms the baseline system by 6.8\% and the single-task model by 2.4\%. We speculate that the gains come from the joint model's ability to help deal with \temprel{none} pairs, since recall scores for \temprel{before} and \temprel{after} increase by 1.5\% and 1.1\% respectively (Table 10  in our  appendix).

\subsection{Event Extraction}

\paragraph{TB-Dense.} Our structured joint model out-performs the CAEVO baseline by 3.5\% and the single-task model by 1.3\%. Improvements on event extraction can be difficult because our single-task model already works quite well with a close-to 89\% F1 score, while the inter-annotator agreement for events in TimeBank documents is merely 87\% \citep{S13-2001}.

\paragraph{MATRES.} %All of our models outperform the baseline system. 
The structured model outperforms the the baseline model and the single-task model by 2.6\% and 0.9\% respectively. However, we observe that the multi-task model has a  slight drop in event extraction performance over the single-task model (86.4\% vs. 86.9\%). This indicates that incorporating relation signals are not particularly helpful for event extraction on MATRES. We speculate that one of the reasons could be the unique event characteristics in MATERS. As we described in Section \ref{sec:data}, all events in MATRES are verbs. It is possible that a more concentrated single-task model works better when events are homogeneous, whereas a multi-task model is more powerful when we have a mixture of event types, e.g., both verbs and nouns as in TB-Dense.

\subsection{Relation Extraction with Gold Events}
\paragraph{TB-Dense.} %Again, previous research based on TB-Dense assumes all events and temporal relations are densely annotated and hence the evaluation uses only gold relation labels. 
There is much prior work on relation extraction based on gold events in TB-Dense. ~\newcite{meng2018context} proposed a neural model with global information that achieved the best results as far as we know.
The improvement of our single-task model over that baseline is mostly attributable to the adoption of BERT embedding. We show that sharing the LSTM layer for both events and relations can help further improve performance of the relation classification task by 2.6\%. For the joint models, since we do not train them on gold events, the evaluation would be meaningless. We simply skip this evaluation.

\paragraph{MATRES.} Both single-task and multi-task models outperform the baseline by nearly 10\%, while the improvement of multi-task over single task is marginal. In MATRES, a relation pair is equivalent to a verb pair, and thus\temprel{} the event prediction task probably does not provide much more information for relation extraction.

\begin{table}
    \small 
 	\centering
 	\setlength{\tabcolsep}{0.5em}
 	\begin{tabular}{|@{ }c@{}|c@{ }|c@{ }|c@{ }|c@{ }|c@{ }|c@{ }|c@{ }|c@{ }|c@{ }|} \hline
 	&\multicolumn{3}{|@{ }c@{ }|}{\textbf{CAEVO}}
 	&\multicolumn{3}{|@{ }c@{ }|}{\textbf{Pipeline Joint}} & \multicolumn{3}{|@{ }c@{ }|}{\textbf{Structure Joint}}\\ \cline{2-10}
 	& \textbf{P}& \textbf{R} & \textbf{F1}&\textbf{P}& \textbf{R} & \textbf{F1}&\textbf{P}& \textbf{R} & \textbf{F1}\\ \hline
 	 \textbf{B}&41.4&19.5&26.5& 59.0& 46.9& 52.3 & 59.8 & 46.9 & 52.6\\
 	 \textbf{A} & 42.1&17.5&24.7& 69.3 & 45.3 & 54.8 & 71.9 &  46.7& 56.6\\
 	 \textbf{I}&50.0&3.6&6.7&-&-&-&-&-&-\\
 	 \textbf{II}&38.5&9.4&15.2&-&-&- &-&-&-\\
 	 \textbf{S}&14.3&4.5&6.9&-&-&-&-&-&-\\
 	 \textbf{V}&44.9&59.4&51.1&45.1&55.0&49.5 &45.9&55.8&50.4\\\hline \hline
 	 \textbf{Avg}&43.8&35.7&39.4&51.5&  45.9& 48.5 & 52.6 & 46.5& \textbf{49.4}\\ \hline
 	\end{tabular}
   	\caption{Model performance breakdown for TB-Dense. ``-'' indicates no predictions were made for that particular label, probably due to the small size of the training sample. \small{\temprel{Before} (\textbf{B}), \temprel{After} (\textbf{A}), \temprel{Includes} (\textbf{I}), \temprel{Is\_Included} (\textbf{II}), \temprel{Simultaneous} (\textbf{S}), \temprel{vague} (\textbf{V})}}
   	\vspace{-0.5cm}
   	\label{tab:tbd-breakdown}
 \end{table}

In Table ~\ref{tab:tbd-breakdown} we further show the breakdown performances for each positive relation on TB-Dense. The breakdown on MATRES is shown in Table 10 in the appendix. \temprel{before}, \temprel{after} and \temprel{vague} are the three dominant label classes in TB-Dense. We observe that the linguistic rule-based model, CAEVO, tends to have a  more evenly spread-out performance, whereas our neural network-based models are more likely to have concentrated predictions due to the imbalance of the training sample across different label classes.

\subsection{Discussion}
\paragraph{Label Imbalance.} One way to mitigate the label imbalance issue is to increase the sample weights for small classes during model training. We investigate the impact of class weights by refitting our single-task model with larger weights on \temprel{includes}, \temprel{is\_included} and \temprel{vague} in the cross-entropy loss.

 \begin{table}
 	\centering
  	\small
 	\setlength{\tabcolsep}{0.5em}
 	\begin{tabular}{|l|c|c|} \hline
 	\textbf{Labels}&\textbf{TB-Dense} & \textbf{MATRES}\\ \hline\hline
 	\temprel{before} & 384 & 417\\
 	\temprel{after} & 274 & 266\\
 	\temprel{includes} & 56 & --\\
 	\temprel{is\_included} & 53 & --\\
 	\temprel{simultaneous} & 22 & 31\\
 	\temprel{vague} & 638 & 113\\
 	\hline
 	\end{tabular}
   	\caption{Label Size Breakdown in the Test Data}
   	\vspace{-0.4cm}
   	\label{tab:class_size}
 \end{table}

\begin{figure}[t]
\centering
\includegraphics[width=0.9\columnwidth]{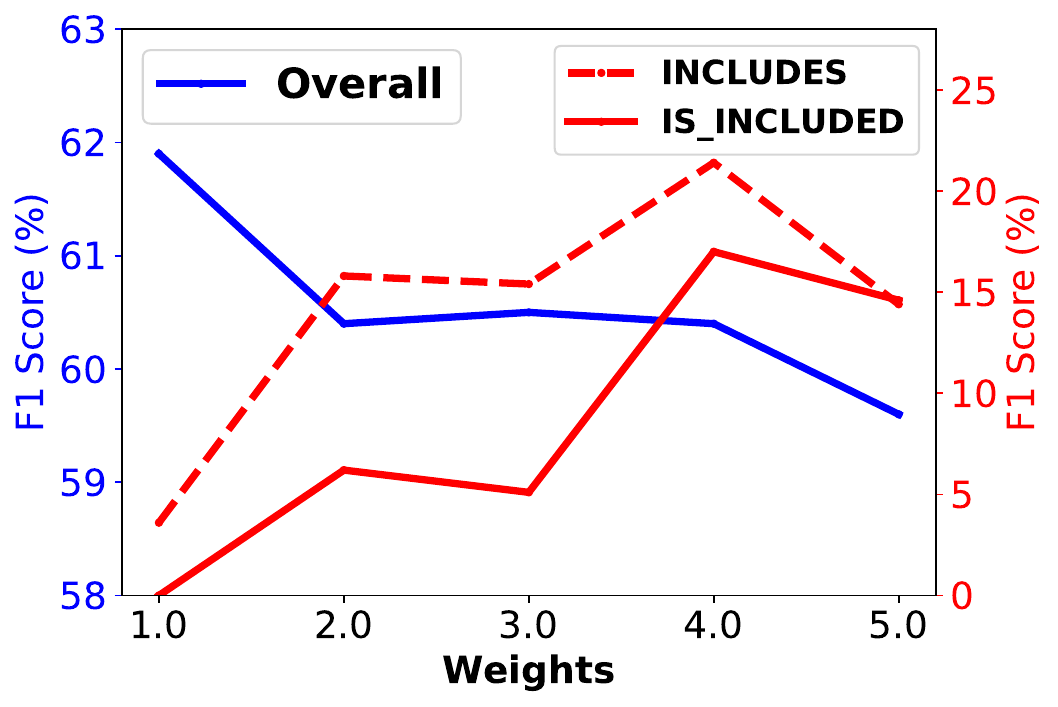}
\caption{
\label{fig:class_weight} Performances from a single-task relation model under different class weights. Left-axis: overall model; right-axis: two minority relations.
}
\vspace{-0.5cm}
\end{figure}

Figure ~\ref{fig:class_weight} shows that increasing class weights up to 4 times can significantly improve \qiangchange{the} F1 scores of \temprel{includes} and \temprel{is\_included} classes 
\qiangchange{with a decrease less than 2\% for the overall F1 score.}
% with less than a 2\% decrease of overall model F1 score. 
Performance of \temprel{includes} and \temprel{is\_included} eventually degrades when class weights are too large. These results seem to suggest that more labels are needed in order to improve the performance on both of these two classes and the overall model. For \temprel{simultaneous}, our model does not make any correct predictions for both TB-DENSE and MATRES by increasing class weight up to 10 times, which implies that \temprel{simultaneous} could be a hard temporal relation to predict in general.

 \begin{table}
 	\centering
 	\small
 	\setlength{\tabcolsep}{0.5em}
 	\begin{tabular}{|l|c|c|} \hline
 	\textbf{Micro-average}&\textbf{TB-Dense} & \textbf{MATRES}\\ \hline\hline
 	No Structure & 48.5 & 58.5\\
 	$+$ $\mathcal{E}\mathcal{R}$ Consistency & 49.4 & 59.5\\
 	$+$ Transitivity& 49.4 & 59.6\\\hline
 	\end{tabular}
   	\caption{Ablation 
   	Study on Global Constraints}
   	\label{tab:ablation}
   	\vspace{-0.5cm}
 \end{table}
 
\paragraph{Global Constraints.} In Table~\ref{tab:ablation} we conduct an  ablation study to understand the contributions from the event-relation prediction consistency constraint and the temporal relation transitivity constraint for the structured joint model. As we can see, the event-relation consistency help\qiangchange{s improve} the F1 scores by 0.9\% and 1\% for TB-Dense and MATRES, \qiangchange{respectively,} but the gain by using transitivity is either non-existing or marginal. We hypothesize two \qiangchange{potential} reasons:
% that might potentially explain this: 
1) We leveraged BERT contextualized embedding as word representation, which could tackle transitivity in the input context; 2) \temprel{none} pairs could make transitivity rule less useful, as positive pairs can be predicted as \temprel{none} and transitivity rule does not apply to \temprel{none} pairs.

\paragraph{Error Analysis.} By comparing gold and predicted labels for events and temporal relations and examining predicted probabilities for events, we identified three major sources of mistakes made by our structured model, as illustrated in Table~\ref{tab:error_types} with examples.
\vspace{-0.2cm}
\paragraph{Type 1.}  Both events in Ex 1 are assigned low scores by the event module ($<< 0.01$). Although the structured joint model is designed to predict events and relations jointly, we leverage the event module to filter out tokens with scores lower than a threshold. Consequently, some true events can be mistakenly predicted as non-events, and the relation pairs including them are automatically assigned \temprel{none}. 
\vspace{-0.2cm}
\paragraph{Type 2.} In Ex 2 the event module assigns high scores to tokens \textbf{happened} (0.97) and \textbf{according} (0.89), but \textbf{according} is not an event. When the structured model makes inference jointly, the decision will weigh heavily towards assigning \textbf{1} (event) to both tokens. With the event-relation consistency constraint, this pair is highly likely to be predicted as having a  positive temporal relation. Nearly all mistakes made in this category follow the same pattern illustrated by this example.
\vspace{-0.2cm}
\paragraph{Type 3.} The existence of \temprel{vague} makes temporal relation prediction challenging as it can be easily confused with other temporal relations, as shown in Ex 3. This challenge is compounded with \temprel{none} in our end-to-end extraction task. 

 \begin{table}
 	\centering
  	\small
 	\setlength{\tabcolsep}{0.5em}
 	\begin{tabular}{|l|r|} \hline
 	\textbf{Type 1}: Event predicted as non-event & 189 pairs\\ \hline
 	\multicolumn{2}{|l|}{\textbf{Ex 1.} What Microsoft gets are developers around the} \\
 	\multicolumn{2}{|l|}{world \textbf{working} on ideas that could potentially \textbf{open} } \\
 	\multicolumn{2}{|l|}{up Kinect for Windows ... } \\\hline \hline
 	\textbf{Type 2}: \temprel{NONE} predicted as true relation & 135 pairs\\ \hline
 	\multicolumn{2}{|l|}{\textbf{Ex 2.} Mr. Netanyahu told Mr. Erdogan that what} \\
 	\multicolumn{2}{|l|}{\textbf{happened} on board the Mavi Marmara was} \\
 	\multicolumn{2}{|l|}{``unintentional'' ... , \textbf{according} to the statement.} \\ \hline \hline
 	\textbf{Type 3}: \temprel{vague} relation & 87 pairs\\ \hline
 	\multicolumn{2}{|l|}{\textbf{Ex 3.} Microsoft \textbf{said} it has identified 3 companies for} \\
 	\multicolumn{2}{|l|}{the China program to run through June. The company} \\
 	\multicolumn{2}{|l|}{\textbf{gives} each participating startup \$ 20,000 to create ...} \\ \hline
 	\end{tabular}
   	\caption{Error Types Based on MATRES Test Data}
   	\label{tab:error_types}
   	\vspace{-0.5cm}
 \end{table} 
 
 Type 1 and Type 2 errors suggest that building a stronger event detection module will be helpful for both event and temporal relation extraction tasks. To improve the performance on \temprel{vague} pairs, we could either build a stronger model that incorporates both contextual information and commonsense knowledge or create datasets with annotations that better separate \temprel{vague} from other positive temporal relations.
\section{Conclusion}
\vspace{-0.2cm}
In this paper we investigate building an end-to-end event temporal relation extraction system. We propose a novel neural structured prediction model with joint representation learning to make predictions on events and relations simultaneously; this can avoid error propagation in previous pipeline systems. Experiments and comparative studies on two benchmark datasets show that the proposed model is effective for end-to-end event temporal relation extraction. Specifically, we improve the performances of previously published systems by 10\% and 6.8\% on the TB-Dense and MATRES datasets, respectively.

Future research can focus on creating more robust structured constraints between events and relations, especially  considering event types, to improve the quality of global assignments using ILP. Since a better event model is generally helpful for relation extraction, another promising direction would be to incorporate multiple datasets to enhance the performance of our event extraction systems.
%and consider fine-grained event types and event-argument structures,
%either in construction relation candidates or filter out non-events and thus reduce \temprel{none} pairs.

\section*{Acknowledgements}
\vspace{-0.2cm}
% We thank the anonymous reviewers for their constructive comments, as well as the members of
% the USC PLUS lab for their early feedbacks. 
This work is supported in part by Contracts W911NF-15-1-0543 and HR0011-18-2-0052 with the US Defense Advanced Research
Projects Agency (DARPA). Approved for Public Release, Distribution Unlimited. The views expressed are those of the authors and do not reflect the official policy or position of the Department of Defense or the U.S. Government.

\clearpage
\bibliography{emnlp-ijcnlp-2019}
\bibliographystyle{acl_natbib}
\end{document}